# Implicit Sensitive Text Summarization based on Data Conveyed by Connectives

Henda Chorfi Ouertani

Information Technology Department
College of Computer and Information Sciences, King Saud University
Riyadh,Saudi Arabia

*Abstract*—**So far and trying to reach human capabilities, research in automatic summarization has been based on hypothesis that are both enabling and limiting. Some of these limitations are: how to take into account and reflect (in the generated summary) the implicit information conveyed in the text, the author intention, the reader intention, the context influence, the general world knowledge…. Thus, if we want machines to mimic human abilities, then they will need access to this same large variety of knowledge. The implicit is affecting the orientation and the argumentation of the text and consequently its summary. Most of Text Summarizers (TS) are processing as compressing the initial data and they necessarily suffer from information loss. TS are focusing on features of the text only, not on what the author intended or why the reader is reading the text. In this paper, we address this problem and we present a system focusing on acquiring knowledge that is implicit. We principally spotlight the implicit information conveyed by the argumentative connectives such as: but, even, yet …. and their effect on the summary.**

## I. INTRODUCTION

Nowadays, text summarization has become widely used on the internet. Users of text summarization are countless. They can be simple internet surfers searching for different news, e-learners looking for specific educational materials or scientists exploring particular publications… Text summarization can help those users identify, in a short time (by reducing a large amount of information in a summary), which documents are most relevant to their needs. But, there is widespread agreement that summarization that reduces a large volume of information to a summary preserving only the most essential items, is a very hard process. Indeed, the human summarization is the process that given a document one tries to *understand, interpret, abstract* it and finally *generate* a new document as its summary [1].

So far and trying to reach human capabilities, research in automatic summarization has been based on hypothesis that are both enabling and limiting. Some of these limitations are: how to take into account and reflect (in the generated summary) the implicit information conveyed in the text, the author intention, the reader intention, the context influence, the general world knowledge ... Thus, If we want machines to mimic human abilities, then they will need access to this same large variety of knowledge [2].

Most of Text Summarizers (TS) are processing as compressing the initial data and they necessarily suffer from information loss. TS are focusing on features of the text only, not on what the author intended or why the reader is reading the text. Thus a TS system must identify important parts and preserve them. In this paper, we will focus on acquiring knowledge that is implicit in the data and how to preserve it when generating the summary. The system we present generate argumentative text based on the implicit stored data conveyed by the "argumentative connectives" such as nevertheless, therefore, but, little, a little... When those connectives appear in sentences, they impose constraints on the argumentative movement. This movement is based on gradual rules of inference denoted by "topoi" [3]

The paper is organized as follows: in section 2, we give an overview of the state of the art on text summarization. Section 3 reports on the theory of Argumentation Within Language (AWL) on which is based our implicit extractor. In section 4, we describe our system architecture. In conclusion, we summarize the contributions of this paper and introduce future research directions.

## II. TEXT SUMMARIZATION

### A. Types of summarizers

Text summarization is now an established field of natural language processing, attracting many researchers and developers. We can distinguish two types of summarizers based on the volume of text to be summarized:

- Single Document Summarization (SDS): If summarization is performed for a single text document then it is called as the single document text summarization
- Mutli Document Summarization (MDS) : If the summary is to be created for multiple text documents then it is called as the multi document text summarization

### B. Summarization techniques

Techniques may vary depending on the summarization type. When considering the Single Document Summarization, we can cite the most important techniques:

- Sentences extracting: This technique relies on trivial features of sentences, such as word frequency, presence of keywords, and sentence position, or a combination of such features [4], [5].

- Identification of the relevant information: permitting to generate a textual summary from the facts that need to be included [6], [7].

However, when dealing with Multi-document summarization, we can talk about

- Extractive summarization: this technique involves assigning scores to some units (e.g. sentences, paragraphs) of the documents and extracting those with highest scores [8].

- Abstractive summarization: this technique usually needs information fusion, sentence compression and reformulation [4].

### III. How connectives are affecting sentence orientation

#### A. Introduction

In order to show the importance of the connective on the orientation of the sentence and on its general meaning, we used LSA tool (http://lsa.colorado.edu/) to compare two apparently same sentences. LSA is a theory and a method for extracting and representing the contextual usage meaning of words by statistical computation. LSA measures of similarity are considered highly correlated with human meaning similarities among words and texts. Moreover, it successfully imitates human word selection and category judgments [9].

Example 1:
Let us consider the two following sentences:

*1) The weather is beautiful but I have to work*
*2) I have to work but the weather is beautiful*

With LSA the two sentences will be represented with the same semantic vectors (fig. 1.) because for LSA the words like I, to, but … are ignored and the word order is not take into account.

| Sentence to Sentence Coherence Comparison Results |||
|---|---|---|
| The submitted texts' sentence to sentence coherence: |||
| **COS** | **SENTENCES** ||
| 1.0 0 | *1:* | The weather is nice but I have to work. |
| | *2:* | I have to work but the weather is nice. |
| **Mean of the Sentence to Sentence Coherence is: 1.00** |||
| **Standard deviation of the Sentence to Sentence is: 0.00** |||

Fig. 1. Comparison of two sentences similarity, Comparison from http://lsa.colorado.edu/

But we agree that the two sentences argue to two different conclusions. So, it is definitely the impact of ignoring the connective *but*.

#### B. Argumentation Within Language Theory

The Argumentation Within Language Theory (AWL) [10] has been concerned with the analysis of the "argumentative articulators" such as nevertheless, therefore, but, little, a little... When those articulators appear in utterances, they impose on constraints on the argumentative movement. This movement is based on gradual rules of inference denoted by "topoi". According to [11] and [12], a topos is an argumentative rule shared by a given community (which need have no more members than the speaker and the hearer). Topoi are the guarantors of the passage from the argument to the conclusion. Topoi are used to license the move from an argument to a conclusion.

A topos (singular of topoi) is:

- Presented as general: in the sense that the speaker implicates that the topos holds for other situations. It is not particular for the situation where it is used.

- Presented as shared: in the sense that the speaker considers that the topos is accepted at least by the audience.

- Gradual.

The canonical form of the topos includes two argumentative scales: the argument (antecedent) and the conclusion (consequent).

Each scale is marked on "plus" or on "minus" from which the next topical forms are concluded:

$$// + P, + Q//,$$
$$// - P, - Q//,$$
$$// + P, - Q// \text{ and }$$
$$// - P, + Q//.$$

If we believe $// + P, + Q//$, we necessarily believe $// - P, - Q//$ and in the same way for $(// + P, - Q// ; // - P, + Q//)$

To illustrate the presentation above, let us consider the utterance

(1) The weather is beautiful but I have to work.

The antecedent uses a topos such as //plus weather is beautiful, plus we want to go out//, the conclusion uses a topos such as //plus I have a work to do, minus I go out//. The use of "but" in the utterance influences its argumentative orientation and the all utterance orientation will be the orientation of the conclusion.

Let us now consider together the two sentences of example1: According to the AWL, the two sentences have opposite argumentative orientations.

Indeed, for the sentence 1, if the antecedent uses topos like //+ beautiful weather, + outing// and the conclusion uses topos like //+ work, - outing// then the presence of "but" imposes that the sentence have the argumentative orientation of the conclusion i.e. "- outing".

However, for the sentence 2, and with the same reasoning, its argumentative orientation is "+ outing"

To end this illustration, we note the importance of "but", in the sense that it imposes the argumentative orientation of the sentence. This importance of connectives was already revealed by different works on Natural Language Process such as in [13] "interclausal connectives carry meaning, they connect textual meanings at both local and global levels and they mark discourse continuity and discontinuity both in the text and as inferred by the reader"

Connectives can shape the actual meaning of the text, they can also serve as efficient markers for instructions in the communicative process established between writer and reader.

After this short outline on the theory of the Argumentation Within Language, in the next section we give a description of the architecture of an Argumentative Single Document Summarizer (ASDS).

## IV. SYSTEM ARCHITECTURE

This section gives an overview of the ASDS architecture and describes the functions of its various components. The global architecture is represented in Figure 1. It is composed of three layers of software : the Data pre-processor, the constraints generator and the summary generator.

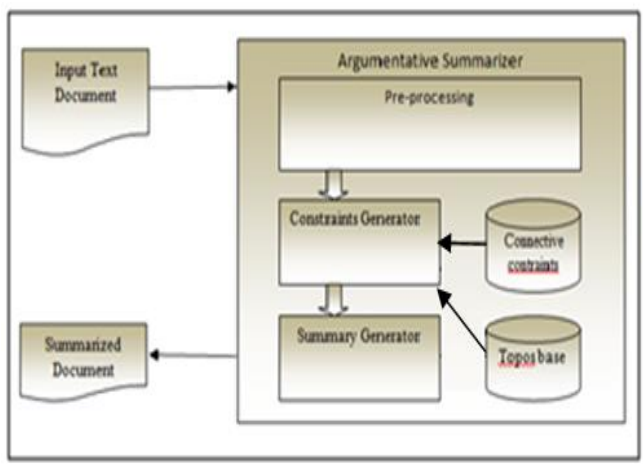

Fig. 2. ASDS Architecture

The pre-processing layer aims at extracting connective elements. ASDS uses GATE [14] a natural language processing system.

The generator constraints layer Generate constraints based on the connectives constraints and the topos base. It permits to annotate the relevant sentences in the text. In our work we consider the sentence as the basic extraction unit. The connective constraints determine the type of argumentative relation between the argument and the conclusion - whether an argument-conclusion relation or argument-anti-argument relation- The topos base is used to link arguments to conclusions. This base allows the comparison of two arguments across scales (since a topos is gradual as discussed above).

We notice that the proposed summarization is focused on single document texts where argumentation takes an important place. The summary generator aims to filter sentence according to the constraints predetermined by the constraints generator. The algorithm below gives the different steps of summary generation :

- Identify all sentences S={Si} of the document d.
- Calculate sentences score with respect to their importance for the overall understanding of the text. This ranking is based on key words and connectives.

Sentences with connectives are weighted contrary to other sentences.

Key words are determined by their frequency in the document.

A Word-Sentence matrix is generated, where the column represents the sentences and the row represents the words. Words with maximum frequency are considered as key words.

Calculate the score for each sentence using a formula using the key words weight and connectives weight :

$$Score(Si) = Cw*Ww$$

|      | S1 | S2 | … | …. | Sn |
|------|----|----|----|----|----|
| W1   |    |    |    |    |    |
| W2   |    |    |    |    |    |
| ..   |    |    |    |    |    |
| …    |    |    |    |    |    |
| Wn   |    |    |    |    |    |
| Ww   |    |    |    |    |    |
| Cw   |    |    |    |    |    |
| Score |   |    |    |    |    |

Where Cw is the weight of connectives and Ww is the weight of key words.

- Rank the sentences in the decreasing order of calculated scores.
- Apply connectives constraints on sentences including connectives to generate conclusions.
- Top ranked sentences and generated conclusions are combined in sequence as document summary.

## V. FUTURE WORK

In the present work, we showed the role of connectives in argumentative texts when dealing with the orientation of the whole text. The analysis of these connectives indicates the existence of specific values intentionally assigned to them by the writer named topoi. As future work, we plan to investigate the topoi base. Many works need to be conducted especially how this base will be initialized and how it will be updated. We would like to continue the implementation of ASDS to apply our approach. Moreover, choosing argumentative texts to be used as input to our system needs further investigation.

## VI. CONCLUSION

In this paper we showed the role of connectives in argumentative texts when dealing with the orientation of the whole text. The analysis of these connectives indicates the existence of specific values intentionally assigned to them by the writer. For example *But* was shown to be functioning in sentence to impose constraints on the conclusion intended by the writer. Some recent trends of investigation support different roles for these connectives in the construction of summaries of argumentative texts. In this context, we present the architecture of ASDS, an Argumentative Single Document Summarizer. ASDS is based on topoi which are gradual rules of inference. Topoi are the guarantors of the passage from the argument to the conclusion.

### ACKNOWLEDGMENT

This research project was supported by a grant from the Research Center of the Center for Female Scientific and Medical Colleges in King Saud University.